\documentclass[letterpaper, 10 pt, journal, twoside]{IEEEtran}
\usepackage{graphicx}
\usepackage{amsmath}
\usepackage{mathbbol}
\usepackage{pifont}
\usepackage{url}
\usepackage{caption}
\usepackage{color}
\usepackage{graphicx}
\usepackage{hyperref}
\usepackage{amsmath}
\usepackage{mathbbol}
\usepackage{pifont}
\usepackage{url}
\usepackage{caption}
\usepackage{color}

\title{
Unleashing the Potential of Mamba: Boosting a LiDAR 3D Sparse Detector by Using  Cross-Model Knowledge Distillation
}

\author{Rui Yu$^{1}$,  Runkai Zhao$^{2}$, Jiagen Li$^{1}$, Qingsong Zhao$^{3}$, Huaicheng Yan$^{1}$,~\IEEEmembership{Senior Member,~IEEE}, Meng Wang$^{1}$

\thanks{Manuscript received: Apr. 4, 2026; Revised: Jul. 8, 2026; Accepted: Aug. 4, 2026. This paper was recommended for publication by Editor Abhinav Valada upon evaluation of the Associate Editor and Reviewers' comments.}
\thanks{Code is available at \href{https://github.com/YuruiAI/FASD}{https://github.com/YuruiAI/FASD}}%
\thanks{$^{1}$ East China University of Science and
Technology,
        {\tt\small \{y80220166, y80220334\}@mail.ecust.edu.cn, \{hcyan, mengwang\}@ecust.edu.cn
}}%
\thanks{$^{2}$ University of Sydney,
        {\tt\small rzha9419@uni.sydney.edu.au}}%
\thanks{$^{3}$ Fudan University,
        {\tt\small qingsongzhao@fudan.edu.cn}}%
}

\begin{document}

\maketitle

\begin{abstract}
The LiDAR 3D object detector that strikes a balance between accuracy and speed is crucial for achieving real-time perception in autonomous driving.
However, many existing LiDAR detection models depend on complex feature transformations, leading to poor real-time performance and high resource consumption, which limits their practical effectiveness.
In this work, we propose a \textbf{\underline{F}}aster LiDAR 3D object detector, a framework that \textbf{\underline{A}}daptively aligns \textbf{\underline{S}}parse voxels to enable efficient heterogeneous knowledge \textbf{\underline{D}}istillation, called \textbf{FASD}.
We aim to distill the Transformer's sequence modeling capability into Mamba models, significantly boosting accuracy through knowledge transfer.
Specifically, we first design the architecture for cross-model knowledge distillation to {{convey}} the global contextual understanding capabilities of the Transformer to Mamba.
Transformer-based teacher model {{employs}} a scale-adaptive attention mechanism to enhance multi-scale fusion.
In contrast, the Mamba-based student model leverages feature alignment through {{spatial alignment}} adapters, supervised with latent space feature and span-head {{logit distribution}}, leading to improved performance and efficiency.
We evaluated FASD on the Waymo and nuScenes datasets, achieving up to a ×2 reduction in FLOPs and a ×4 reduction in memory consumption, while improving baseline performance by 1–2 {{percentage points}} and maintaining efficient deployment.


\end{abstract}  
\section{INTRODUCTION}
\IEEEPARstart{L}{iDAR} 3D object detection is vital to provide 3D object localization and geometric characterization for autonomous driving and robotics navigation \cite{pointpillar, voxelnet, swift}. 
Due to its laser pulse mechanism, LiDAR provides a global receptive field, where foreground points offer rich geometric cues for target characterization.
This characteristic facilitates understanding of the overall scene geometry by leveraging informative sparse features.
Meanwhile, incorporating global context and spatial information \cite{swformer,dsvt} helps the model better understand the interactions between voxel features.

As demonstrated in previous studies \cite{mppnet,msf, listm}, the construction of effective long dependencies helps detection models understand contextual associations.
Transformer-based LiDAR object detectors \cite{sst, swformer, dsvt} utilize the attention mechanism and positional embeddings to encode global contextual understanding and local spatial information.
This token-wise interaction of neighboring voxel features enhances object representation and scene understanding.

\begin{figure}
    \centering
    \includegraphics[width=0.45\textwidth]{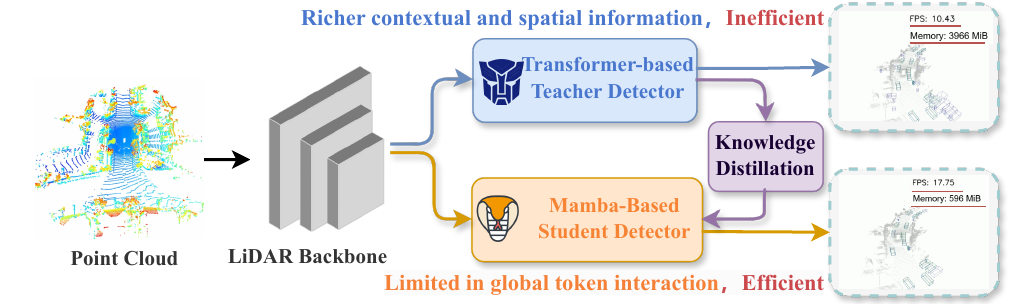}
    \caption{\textbf{Overview of the Proposed FASD Framework}. We {{transfer}} the {{Transformer’s contextual and spatial knowledge to the efficient Mamba architecture, improving detection performance while preserving inference efficiency.}}}
    \label{fig1}
    \vspace{-0.2cm}
\end{figure}

Nevertheless, the Transformer-based model suffers from the high computational demand caused by computing the query-key attention matrix with quadratic complexity \cite{longformer}, which limits its application in real-world autonomous driving.
Although some methods \cite{voxelnext,safdnet} improve efficiency by bypassing densification and using sparse representations, they encounter performance bottlenecks.
Meanwhile, methods like Linformer \cite{linformer} and Preformer \cite{preformer} reduce computational complexity through approximately linear attention mechanisms.
However, Mamba \cite{mamba} takes a different approach by utilizing linear-time sequence modeling with selective scan strategies and efficient hardware-aware algorithms to optimize token selection and data flow.
By leveraging sequential data and sparse voxel structures, Mamba demonstrates superior performance over Transformer \cite{attention} in terms of FLOPs. 
Unfortunately, unlike Transformer, Mamba processes sequence data by recursively aggregating visual information into a latent vector, without explicitly modeling contextual cues and positional relationships.
Thus, by distilling knowledge from Transformer, we retain the efficient Mamba model, enhancing position sensitivity and global context without increasing complexity.

\begin{figure}
    \centering
    \includegraphics[width=0.45\textwidth]{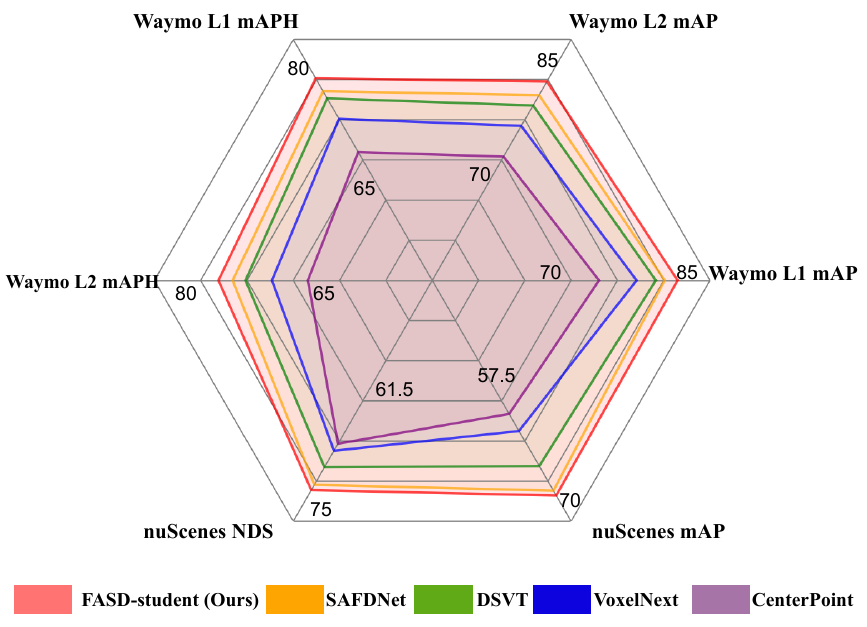}
    \caption{\textbf{Performance comparisons of various existing LiDAR 3D detection models.} Through our proposed FASD framework, the Mamba-based student model {{achieves the competitive performances across the Waymo and nuScenes validation sets.}} For simplicity, we omit the performance of the Transformer-based teacher model here.}
    \label{fig2}
    \vspace{-0.15cm}
\end{figure}

In order to balance the computational efficiency and contextual understanding of LiDAR detectors in real-time environmental sensing,
we propose a \textbf{F}aster LiDAR 3D object detection framework by \textbf{A}daptive aligning \textbf{S}parse voxel to enable heterogeneous knowledge \textbf{D}istillation, namely \textbf{FASD}, as shown in Fig.~\ref{fig1}.
Overall, we design a training paradigm for heterogeneous model distillation from Transformer to Mamba, integrating dynamic voxel grouping and diffusion module to enhance contextual and spatial information through sequence interactions.
For the teacher model, we enhance long-sequence modeling using an adaptive attention mechanism to fuse multi-scale features, effectively capturing global context.
For the student model, we replace the baseline with Mamba for improved efficiency and adopt local-global grouping to enable long-range modeling.
During cross-model distillation, we perform spatial alignment using an adapter to mitigate misalignment.
Additionally, Span-KD projects cross-model features into a unified logit space and enforces probability distribution consistency, thereby enhancing global contextual modeling and spatial geometry understanding.
Fig \ref{fig2} illustrated the performance, {{where our model achieves competitive performance compared with existing baselines.}}

The main contributions can be summarized as follows:
\begin{itemize}
\item To balance detection performance and efficiency, we design {{a}} cross-model distillation paradigm, {{enabling effective knowledge transfer}} from Transformer to Mamba.
\item We propose dynamic voxel group module and adaptive attention mechanism to convert voxels into sequences, enabling scale-adaptive extraction, building a teacher model with better spatial-contextual understanding.
\item We design heterogeneous distillation in latent and logit spaces via {{an}} alignment adapter, enhancing the Mamba-based student with global context and geometric awareness without increasing operational complexity.
\item {We conduct experiments on the Waymo and nuScenes datasets to validate our method, demonstrating a \textbf{1–2\%} performance improvement over the baseline while achieving \textbf{×2} lower FLOPs, \textbf{×4} lower memory consumption, and real-time inference in real-world applications.}
\end{itemize}  
\section{RELATED WORKS}

\subsection{{LiDAR} 3D Object Detector}
As a vital sensor in autonomous driving, LiDAR provides accurate geometric representations of objects from a bird's-eye view (BEV) perspective, enhancing spatial understanding and characterization.
Seminal works \cite{pointpillar, voxelnet} make substantial contributions to the field, each proposing unique methods for transforming point clouds into a latent space.
Subsequent works \cite{sst, fsd} further improve accuracy and efficiency through channel-wise transformers and sparse voxel attention.
To tackle the challenges of sparse target features and the high computational complexity of submanifold and regular sparse convolutions, Wang et al. \cite{dsvt} introduce the Dynamic Sparse Voxel Transformer, which processes sparse local regions in parallel.
Two-stage LiDAR detectors \cite{detzero, pvrcnn++} use coarse 3D proposals and keypoints as priors, while motion-based detectors \cite{mppnet, msf, listm} excel in high-precision offline detection by processing point clouds and trajectories from cross-frame with Transformer-based spatiotemporal encoding.
Mamba-based models \cite{unimamba,lion,voxelmamba} employ a more efficient encoder architecture, leading to improved performance and efficiency.
These works provide the foundation for our method, while we focus on developing an efficient model for capturing both scene context and spatial details.

\subsection{{Sparse Object Representation}}
To reduce complexity and create lightweight representations, sparsebev \cite{sparsebev} replaces BEV features with sparse voxel or pillar queries for efficient environmental characterization.
In LiDAR 3D object detection, SST \cite{sst} improves efficiency by targeting unique voxels with sparse region attention, while FSD \cite{fsd} enhances object spatial information using Instance Point Grouping and Sparse Instance Recognition. 
Meanwhile, Chen et al. \cite{voxelnext} introduce a full sparse voxel detector and uses query voxels for efficient prediction and tracking.
Sun et al. \cite{swformer} use a sparse Transformer with bucketing-based window partitioning to achieve high accuracy. 
Zhang et al. \cite{safdnet} introduce a hierarchical encoder-decoder with sparse feature diffusion for improved 3D object detection.
In contrast, we utilize sparse characterization and enable modeling of voxel features using efficient Mamba models.

\subsection{Knowledge Distillation}
Knowledge distillation (KD) aims to enable a compact student model to mimic a larger teacher, thereby inheriting its embedded knowledge \cite{OFAKD, SUMMER, fdta}.
Subsequent improvements in logits-based KD include incorporating structural information \cite{rkd} and KL divergence loss \cite{distkd} to bridge the capacity gap.
Yu et al. \cite{distilldrive} propose a cross-architecture method that aligns intermediate features into a logit space to distill knowledge from heterogeneous models.
SparseKD \cite{sparsekd} distills knowledge into a compact student model with reduced depth, width, and input size, achieving high accuracy with less complexity.
Furthermore, {{CaKDP}} \cite{CaKDP} leverage category-aware distillation and pruning for efficient 3D object detection.
Zhao et al. \cite{lanecmkt} propose a dual-path mechanism to transfer 3D cues from a LiDAR model to an image model.
Therefore, we distill knowledge from Transformers to Mamba by directly comparing heterogeneous features and logit distributions, {{thereby enhancing the resource-efficient Mamba with global contextual understanding and spatial information}}.

\begin{figure*}[t] 
    \centering
    \includegraphics[width=1.0\textwidth]{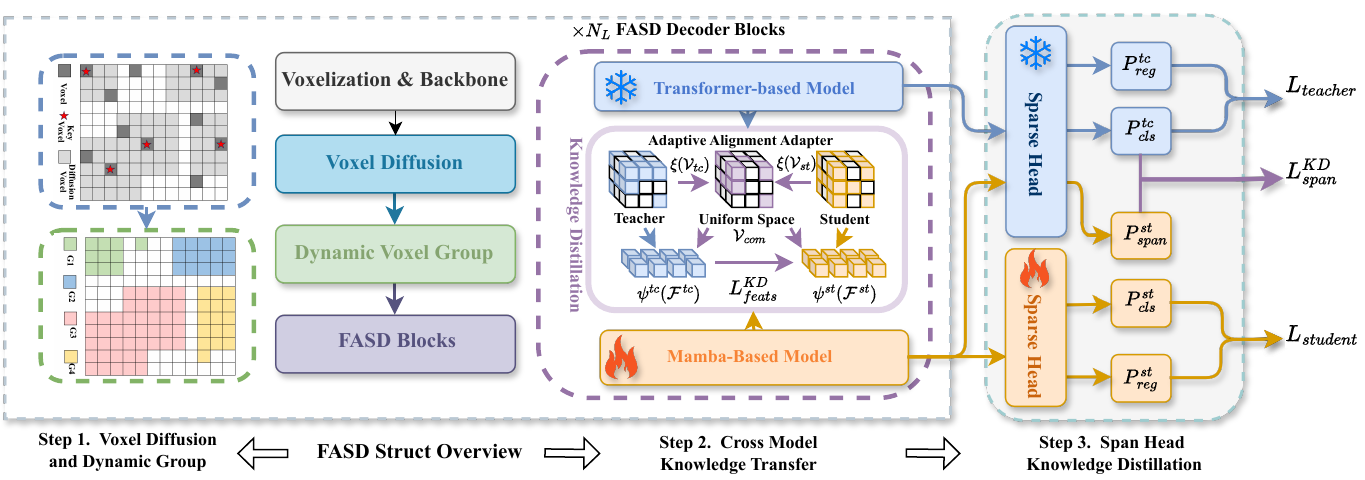}  
    \caption{\textbf{Overview of our proposed FASD pipeline.} 
    Initially, all voxel features are passed through the dynamic voxel group to obtain serialized representations \(\mathcal{F} \in\mathbb{R}^{G\times S \times C}\).
    Subsequently, FASD can be divided into the Transformer-based Teacher Model, the Mamba-based Student model, and the Knowledge Distillation.
    The frozen teacher model guides the student by providing comprehensive supervision over both global contextual information and local spatial features.
    While we use the adapter and span head to better align heterogeneous features, facilitating the distillation of key voxels.}
    \label{fig3}
    \vspace{-0.15cm}
\end{figure*} 
\section{Methodology}

\subsection{Overview}
As illustrated in Fig. \ref{fig3}, we leverage a Transformer model with global contextual understanding to transfer knowledge to a resource-efficient Mamba-based student model.
To begin with, we introduce the model design, including a general-purpose module, the Transformer-based teacher model, and the Mamba-based student model.
Subsequently, we propose faster adaptive sparse distillation to facilitate heterogeneous model distillation from Transformer to Mamba.
Finally, we introduce the training object, which employs co-supervision through teacher guidance and ground truth to enhance the performance of the suboptimal Mamba-based model.

\subsection{Model Design}
Initially, we introduce general-purpose modules, including a sparse backbone, voxel diffusion, and dynamic voxel grouping, to enhance sequence feature representation.
For the teacher model, we achieve multi-scale feature fusion through an adaptive attention mechanisms, which guide the Mamba-based student model in learning global context information.

\noindent\textbf{Sparse Backbone and Neck.}
The sparse backbone employs Submanifold Convolution for implicit feature characterization and Sparse Convolution for downsampling. 
Meanwhile, the neck component encodes and decodes features to facilitate multi-scale fusion.
This design enables the model to capture long-range dependencies by promoting information exchange between spatially disconnected elements.

\noindent\textbf{Voxel Diffusion.}
As described in previous work \cite{swformer}, voxel diffusion effectively densifies foreground features by incorporating central voxel segmentation, thereby enhancing the performance of the fully sparse detector.
Similarly, we perform diffusion on sparse voxels, utilizing the segmentation \( p^{seg}_{i} \) to identify and diffuse the majority of foreground voxels.
The corresponding segmentation label \( gt_{i} \) is defined in SAFDNet \cite{safdnet}.
This segmentation model is trained alongside the detector, with \( N_{V} \) representing the number of valid voxels.
\begin{align}
L_{seg}=\frac{1}{N_{V}}\sum_{i=1}^{N_{V}}L_{focal}(p^{seg}_{i},gt_{i}).
\end{align}

\noindent\textbf{Dynamic Voxel Group.}
We first partition the voxel \(\mathcal{V}=\{(x_{i},y_{i},z_{i});f_{i}\}_{i=1}^{N_{V}}\) dynamics into \(G\) clusters, with each cluster comprising a fixed number of \(S\) subsets. 
The computational logic follows DSVT \cite{dsvt}, \(\mathcal{Q}_{i}\) denotes the \(i\)-th set partition, and \(\mathcal{D}\) is the voxel ID, yielding the serialized feature \(\mathcal{F} \in\mathbb{R}^{G\times S\times C}\).
Unlike the original design that employs short sequences, we construct longer sequences \(S\) via local-global grouping to capture long-range dependencies, followed by local window refinement to preserve local geometry.
The specific calculation formula is as follows:
\begin{align}
\mathcal{F}_{i}=\mathrm{INDEX}(\mathcal{V},\mathcal{Q}_{i},\mathcal{D}).
\end{align}

\noindent\textbf{Transformer-based Teacher Model.}
Training a dedicated teacher model enhances contextual modeling and facilitates heterogeneous knowledge distillation.
First, sequence feature mapping is performed using linear layers $FC$, where \(\mathcal{P}\) denotes the learnable positional encoding, enabling global feature interaction across groups of different sizes.
\begin{align}
Q,K,V=FC_{Q}(\mathcal{F}+\mathcal{P}), FC_{K}(\mathcal{F}+\mathcal{P}),FC_{V}(\mathcal{F}).
\end{align}

While vanilla multi-head attention provides a global receptive field, it lacks effective local multi-scale feature aggregation.
To address this, we propose adaptive attention mechanism, which learns multi-scale receptive fields guided by learnable queries.
Specifically, we first compute the euclidean distances \( \mathbf{D} \in \mathbb{R}^{N_{Q} \times N_{Q}} \) for all queries within each local group, where \( N_{Q} \) is the number of queries.
Subsequently, the receptive field controller \(\gamma\) is generated via a linear transformation $FC_{\gamma}$ from the query feature, adapting dynamically to each of the \(M\) attention heads.
\begin{align}
\gamma=\{\gamma_{1}, \gamma_{2}, ..., \gamma_{M}\} = FC_{\gamma}(Q).
\end{align}

Thus, each head's receptive field can be tailored to {{capture}} different context scales using the calculated \(\gamma\) and \(\mathbf{D}\).
The specific Adaptive Attention $AdaAttn$ is defined as follows:
\begin{align}
\text{AdaAttn}(Q,K, V ) = \text{Softmax}\left(\frac{QK^{\top}}{\sqrt{d_{k}}} - \gamma \cdot \mathbf{D}\right) V.
\end{align}

 \begin{figure}
    \centering
    \includegraphics[width=0.45\textwidth]{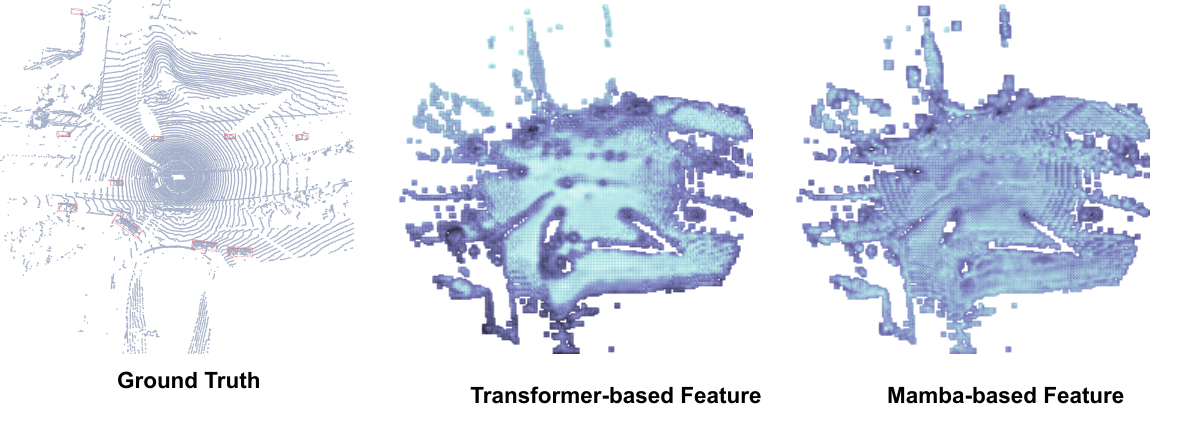}
    \caption{
    Visualization of heterogeneous features shows Transformers capture clearer global context and geometry information, highlighting the need for distillation in Mamba.}
    \label{fig4}
    \vspace{-0.2in}
\end{figure}

As \(\gamma\) increases, attention weights for distant {{tokens}} decrease, shifting focus from a global receptive field to central targets.
Different \(\gamma\) values for each head enable cross-scale feature fusion and enhance global context understanding.

\noindent\textbf{Mamba-based Student Model.}
While Transformers suffer from repeated attention computation, Mamba addresses this issue with an efficient State Space Model (SSM):

\begin{align}
h'(t) &= \overline{A}h(t) + \overline{B}x(t), \\
y(t) &= Ch(t),
\end{align}

\noindent where matrices \(\overline{A}\) and \(\overline{B}\) control the influence of the current state and inputs on state changes, respectively, while matrices \(C\) projects the hidden state to the output. 
Specifically, \(C\) is generated via a linear transformation, while  \(\overline{A}\) and \(\overline{B}\) are learnable parameters, defined as follows:
\begin{align}
\overline{A} = \exp^{\Delta A}, \quad \overline{B} = \frac{(\exp^{\Delta A}-I) \times\Delta B}{\Delta A}.
\end{align}

As described in the dynamic voxel group, non-empty voxels are first converted into sequence representations \(\mathcal{F}\).
These sequence features are progressively processed from long to short sequences within the SSM framework, facilitating hierarchical feature modeling from global context to local spatial details.
We then employ distillation to transfer the spatial and global context features from the Transformer-based teacher model to the Mamba-based student model.

\subsection{Faster Adaptive Sparse Distillation}
\label{section.C}
As shown in Fig. \ref{fig4}, the Transformer produces features that are closer to the target distribution and more discriminative, demonstrating superior capabilities in modeling global context and local spatial interactions.
Due to its high computational overhead, we replace it with the lightweight Mamba, which preserves long-range contextual modeling capability.

\noindent\textbf{Alignment Adapter.} 
As shown in Fig. \ref{fig3}, we dynamically divide the voxel into sequential features and employ Transformer and Mamba models for scene modeling.
However, the difference in key voxel segmentation between the two models causes inconsistent diffusion results, leading to significant discrepancies in the distilled voxel indices.
Meanwhile, existing distillation schemes focus on isomorphic features and neglect the direct distillation of unstructured sparse voxel features.
To address this, we devise an alignment adapter for accurate sparse voxel distillation.
Given voxel features \( \mathcal{F}_{tc} / \mathcal{F}_{st} \)and voxel coordinates \( \mathcal{V}_{tc} / \mathcal{V}_{st} \) from the teacher and student models, respectively, the mapping function \(\xi\) encodes voxel coordinates into a unified representation to identify the shared voxel set \( \mathcal{V}_{com} \).
\begin{align}
\mathcal{V}_{com} &= \xi(\mathcal{V}_{tc}) \cap \xi(\mathcal{V}_{st}), \\
\xi(\mathcal{V}_{i}) &= \mathcal{V}_{x_i}
+ \mathcal{V}_{y_i} \times 10^{4}
+ \mathcal{V}_{z_i} \times 10^{8}.
\end{align}
where \(\xi(\cdot)\) maps 3D voxel coordinates into a unified representation space, enabling efficient identification of the shared voxel set between the teacher and student models.

As shown in Fig. \ref{fig3} (Step 2), the shared voxel set \( V_{com}\)  is retrieved from the unified representation space, and the alignment adapter \(\psi\) extracts the corresponding voxel features from both models.
Consequently, the distillation process focuses on shared key voxels, avoiding the interference caused by misaligned features.
\begin{align}
\psi_i=
\delta\!\left(\xi(V_i)\in V_{com}\right)F_i,
\end{align}
where \( \delta(.) \) denotes the proposed adaptive spatial alignment operator, which identifies the shared voxel correspondences in \( V_{com} \) and selectively preserves spatially aligned voxel features for accurate cross-model knowledge distillation.

\noindent\textbf{Shallow Feature Transfer.} 
Since our approach involves cross-architecture distillation, {{feature-level knowledge distillation is employed to guide the student model}}.
It enables the student model to effectively mimic the teacher model's backbone features.
For the shallow features, the mean squared error between the student and teacher models is calculated as:
\begin{align}
L^{KD}_{shallow}=\sum_{i=1}^{G}\sum_{j=1}^{S}||\mathcal{F}_{i,j}^{st}-\mathcal{F}^{tc}_{i,j}||_{2}.
\end{align}

Here, \( i \) and \( j \) denote the group and sequence indices, respectively, while  \( \mathcal{F}^{tc} \) and \( \mathcal{F}^{st} \) represent the token features of the teacher and student models, respectively, facilitating feature alignment across heterogeneous architectures.

\begin{table*}[!t]
\centering
\caption{{\textbf{Quantitative comparisons on Waymo val set}, where teacher and student refer to transformer and mamba blocks.}}
\renewcommand{\arraystretch}{1.1}
\resizebox{1.0\linewidth}{!}{
\begin{tabular}{@{\extracolsep{\fill}}l|cc|cc|cc|cc}
\hline
\textbf{Model} & \multicolumn{2}{c|}{\textbf{ALL (mAP/mAPH)$\uparrow$}} 
& \multicolumn{2}{c|}{\textbf{Vehicle (AP/APH)$\uparrow$}} & \multicolumn{2}{c|}{\textbf{Pedestrian (AP/APH)$\uparrow$}} & \multicolumn{2}{c}{\textbf{Cyclist (AP/APH)$\uparrow$}} \\
 & \textbf{L1} & \textbf{L2} & \textbf{L1} & \textbf{L2} & \textbf{L1} & \textbf{L2} & \textbf{L1} & \textbf{L2} \\
\hline
{SparseKD \cite{sparsekd}} & {72.49 / 69.87} & {66.23 / 63.65} & {72.46 / 71.92} & {64.29 / 63.69} &{74.33 / 68.01} & {66.17 / 60.45} & {70.92 / 69.63} & {68.23 / 67.05} \\
{CaKDP \cite{CaKDP}} & {74.25 / 71.43} & {68.02 / 65.61} &
{73.98 / 73.46} & {65.99 / 65.39} & {76.02 / 69.65} & {67.93 / 62.17} & {72.65 / 71.06} & {69.97 / 68.68} \\
DSVT \cite{dsvt}  & 79.02 / 76.62 & 72.96 / 70.56 & 78.63 / 78.12  & 70.84 / 70.32 & 82.42 / 76.89 & 74.97/ 69.27 & 76.03/ 74.85 & 73.07 / 72.11 \\
SAFDNet \cite{safdnet}  & 80.00 / 77.94 & 73.82 / 71.88 & 79.31 / 78.86  & 71.26 / 70.85 & 83.74 / 79.01 & 76.12/ 71.60 & 76.92/ 75.97 & 74.10 / 73.19 \\
{LION \cite{lion}} & {80.58 / 78.65} & {74.35 / 72.38} & {79.55 / 79.13} & {71.12 / 70.75} & {84.02 / 79.57} & {76.58 / 71.98} & {78.17 / 77.26} & {75.37 / 74.43} \\
VoxelMamba \cite{voxelmamba} & 80.99 / 79.01 & 74.91 / 72.95 & 80.07 / 79.67  & 72.04 / 71.65 & 84.49 / 79.95 & 77.06/ 72.64 & 78.41/ 77.43 & 75.61 / 74.57 \\
\hline
FASD(Teacher) &  \underline{81.22} / \underline{79.25} &  75.23 / \mdseries\underline{73.28}  & \bfseries 80.54 / \bfseries 79.97 & \bfseries 72.62 / \bfseries 71.99 & \underline{84.83} / \bfseries 80.46 & \mdseries\underline{77.21} / \bfseries 72.94 & \underline {78.69} / \underline{77.81} & \underline{75.86} / \underline{74.91} \\
FASD(Student) & \bfseries{81.61} / 79.41 & \bfseries{75.51} / \bfseries 73.51 & \underline{80.43} / \underline{79.93} & \underline{72.57} / \underline{71.91} & \bfseries 84.94 / \mdseries\underline{80.29} & \bfseries 77.35 / \mdseries\underline{72.85} & \bfseries 79.45 / \bfseries 78.51 & \bfseries 76.63 / \bfseries 75.78 \\
\hline
\end{tabular}
}
\label{tab:1}
\vspace{-0.cm}
\end{table*}

\noindent\textbf{Deep Feature Transfer.} 
Nevertheless, voxel diffusion may introduce feature misalignment due to the inconsistent segmentation quality between the teacher and student models.
To alleviate this issue, the proposed alignment adapter aligns non-empty voxel features and extracts their shared features (shown in purple in Fig.~\ref{fig3}), enabling distillation to focus on key voxels while suppressing unmatched ones.
Distillation is then achieved based on the feature differences within the same geometric space as follows:
\begin{align}
L^{KD}_{deep} &= \sum_{i=1}^{G}\sum_{j=1}^{S}||\psi^{tc}(\mathcal{F}_{i,j}^{tc})-\psi^{st}(\mathcal{F}_{i,j}^{st})||_{2}.
\end{align}

\noindent\begin{table}
\centering
\caption{\textbf{{{Quantitative}} comparisons on nuScenes val set}.}
\setlength{\tabcolsep}{1.0mm}
\renewcommand{\arraystretch}{1.2}
\resizebox{1.0\linewidth}{!}{
\begin{tabular}{@{}l|ccccccc@{}}
\hline
\textbf{Model} & \textbf{NDS$\uparrow$} & \textbf{mAP$\uparrow$} & \textbf{mATE$\downarrow$} & \textbf{mASE$\downarrow$} & \textbf{mAOE$\downarrow$} & \textbf{mAVE$\downarrow$} & \textbf{mAAE$\downarrow$} \\
\hline
CenterPoint \cite{centerpoint} & 66.29 & 58.77 & 0.2919 & 0.2566 & 0.3692 & 0.2081 & 0.1837 \\
VoxelNext \cite{voxelnext} & 67.09 & 60.55 & 0.3023 & 0.2526 & 0.3701 & 0.2087 & 0.1851 \\
DSVT \cite{dsvt}  & 68.94 & 64.22 & 0.2877 & 0.2611 & 0.3701 & 0.2087 & 0.1851 \\
VoxelMamba \cite{voxelmamba}  & \underline{71.48} & \bfseries{67.54} & \underline{0.2785} & \bfseries{0.2583} & 0.2584 & \underline{0.2687} & 0.1899 \\
\hline
FASD(Teacher) &  70.53 &  66.58 &  {0.2732} & 0.2540 & 0.3024 & 0.2777 & \bfseries 0.1821 \\
FASD(Student) & \bfseries 71.98 & \underline{66.98} & \bfseries{0.2732} & \bfseries 0.2512 & \underline{0.2657} & \bfseries{0.2620} & \underline{0.1830} \\
\hline
\end{tabular}
}
\label{tab:2}
\vspace{-0.25cm}
\end{table}

\noindent\textbf{Span Head Knowledge Distillation.} 
However, feature-level distillation mainly supervises intermediate representations in the latent space.
Instead of directly aligning the teacher and student logits, we apply Span-KD in a unified logit space.
As illustrated in Fig. \ref{fig3}, the aligned features $\mathcal{F}'^{tc}$ and $\mathcal{F}'^{st}$ are both fed into the frozen teacher head $head$, which projects them into a shared probability space to facilitate contextual knowledge transfer:
\begin{align}
p^{st}_{span}=head(F'^{st}), p^{tc}_{cls}=head(F'^{tc}). \label{eq:head}
\end{align}

To further transfer the semantic knowledge of the teacher model, we employ KL divergence to align the output probability distributions of the teacher and student models:
\begin{align}
L^{KD}_{span}
=
\sum_{i}
p^{tc}_{cls}(i)
\log
\frac{p^{tc}_{cls}(i)}
{p^{st}_{span}(i)}.
\end{align}

The frozen teacher head acts as a fixed projector, providing stable supervision in the shared prediction space while {{preventing}} gradient propagation to the prediction head.

Therefore, the overall distillation loss is formulated as a weighted combination of the shallow, deep, and span-head distillation losses:
\begin{align}
L_{KD} =
\lambda_{1} L^{KD}_{\text{shallow}}
+\lambda_{2} L^{KD}_{\text{deep}}
+\lambda_{3} L^{KD}_{\text{span}}
\label{eq:LKD}
\end{align}

\subsection{Training Object}
\label{section.D}
Unlike methods \cite{centerpoint}, which project 3D voxel features into BEV features and process them through multi-task heads for classification and regression, our approach {{adopts}} a sparse voxel head \cite{voxelnext}.
This allows for direct classification and regression on sparse voxel features, enabling more efficient target characterization through spatial index assignment.
To better supervise the student model, the predictions are constrained by {{the corresponding ground-truth heatmap and bounding boxes}}.
Specifically, $L_{reg}$ and $L_{cls}$ are computed {{following}} CenterPoint \cite{centerpoint}.

\begin{table}[h]
\centering
\caption{{\textbf{Accuracy–Efficiency Analysis}} on Waymo under Identical voxelization and sparse convolution settings.}
\fontsize{5}{5}\selectfont
\renewcommand{\arraystretch}{1.2}
\resizebox{1.0\linewidth}{!}{
\begin{tabular}{@{}l|cccc@{}}
\hline
\textbf{Model} & \textbf{Param} & \textbf{Flops} & \textbf{L2 mAP} & \textbf{L2 mAPH} \\
\hline
CenterPoint \cite{centerpoint} & 7.8M & 114.8G & 68.07 &  65.66 \\
SparseKD \cite{sparsekd} & \bfseries 4.0M & \bfseries 47.8G & - & 65.75 \\
CaKDP \cite{CaKDP} & 7.8M & 116.2G & 69.74 & 67.59 \\
\hline
FASD(Teacher) & 9.1M & 118.3G &  \underline{75.23} & \underline{73.28} \\
FASD(Student) & \underline{6.1M} & \underline{53.1G} & \bfseries 75.51 & \bfseries 73.51 \\
\hline
\end{tabular}
}
\label{tab:3}
\vspace{-0.25cm}
\end{table}

\begin{table}[h]
\centering
\caption{\textbf{Computational efficiency analysis} for LiDAR 3D detectors on a real-world LiDAR point cloud dataset.}
\fontsize{5}{5}\selectfont
\renewcommand{\arraystretch}{1.2}
\resizebox{1.0\linewidth}{!}{
\begin{tabular}{@{}l|cccc@{}}
\hline
\textbf{Model} & \textbf{Param} & \textbf{Flops} & \textbf{Memory} & \textbf{FPS} \\
\hline
CenterPoint \cite{centerpoint} & 7.75M & \bfseries 48.5G & 2360 MiB & \bfseries 21.58 \\
VoxelNext \cite{voxelnext} & 19.05M & 57.1G & 1970 MiB & 18.12 \\
SAFDNet \cite{safdnet} & 9.87M & 73.5G & 2070 MiB & 16.84 \\
DSVT \cite{dsvt} & 8.65M & 110.2G & 3966 MiB & 9.58 \\
VoxelMamba \cite{voxelmamba} & \underline{6.73M} & 58.5G & \underline{653 MiB} & 11.27 \\
\hline
FASD(Teacher) & 9.12M & 121.7G &  4085 MiB & 9.15 \\
FASD(Student) & \bfseries 6.13M & \underline{54.7G} & \bfseries 546 MiB & \underline{19.46} \\
\hline
\end{tabular}
}
\label{tab:4}
\vspace{-0.25cm}
\end{table}

\section{Experiment}
\label{exp}

\subsection{Dataset and Metrics}

The Waymo Open dataset \cite{waymodataset} is a {{widely recognized}} benchmark for autonomous driving and environmental perception.
It consists of 1,150 point cloud sequences, with over 200,000 frames in total.
Performance is evaluated using mean Average Precision (mAP) and its heading-weighted variant (mAPH).
Results are reported for LEVEL 1 (L1, easy only) and LEVEL 2 (L2, easy and hard) difficulty levels, covering vehicles, pedestrians, and cyclists.

The nuScenes dataset \cite{nuscenes} is a widely used benchmark for autonomous driving, featuring diverse driving scenarios and comprehensive evaluation metrics.
Performance is evaluated using mean Average Precision (mAP) {{across}} four center distance thresholds and five true positive metrics, including ATE, ASE, AOE, AVE, and AAE.
The nuScenes Detection Score (NDS) further summarizes the overall detection performance by combining mAP with these error metrics.

Finally, to evaluate real-world {{deployment}} efficiency and performance, we collected LiDAR data using the Livox Mid-360 and constructed a real-world evaluation dataset.

\subsection{Experimental Settings }

\begin{figure*}[t]  
    \centering
    \includegraphics[width=1.0\textwidth]{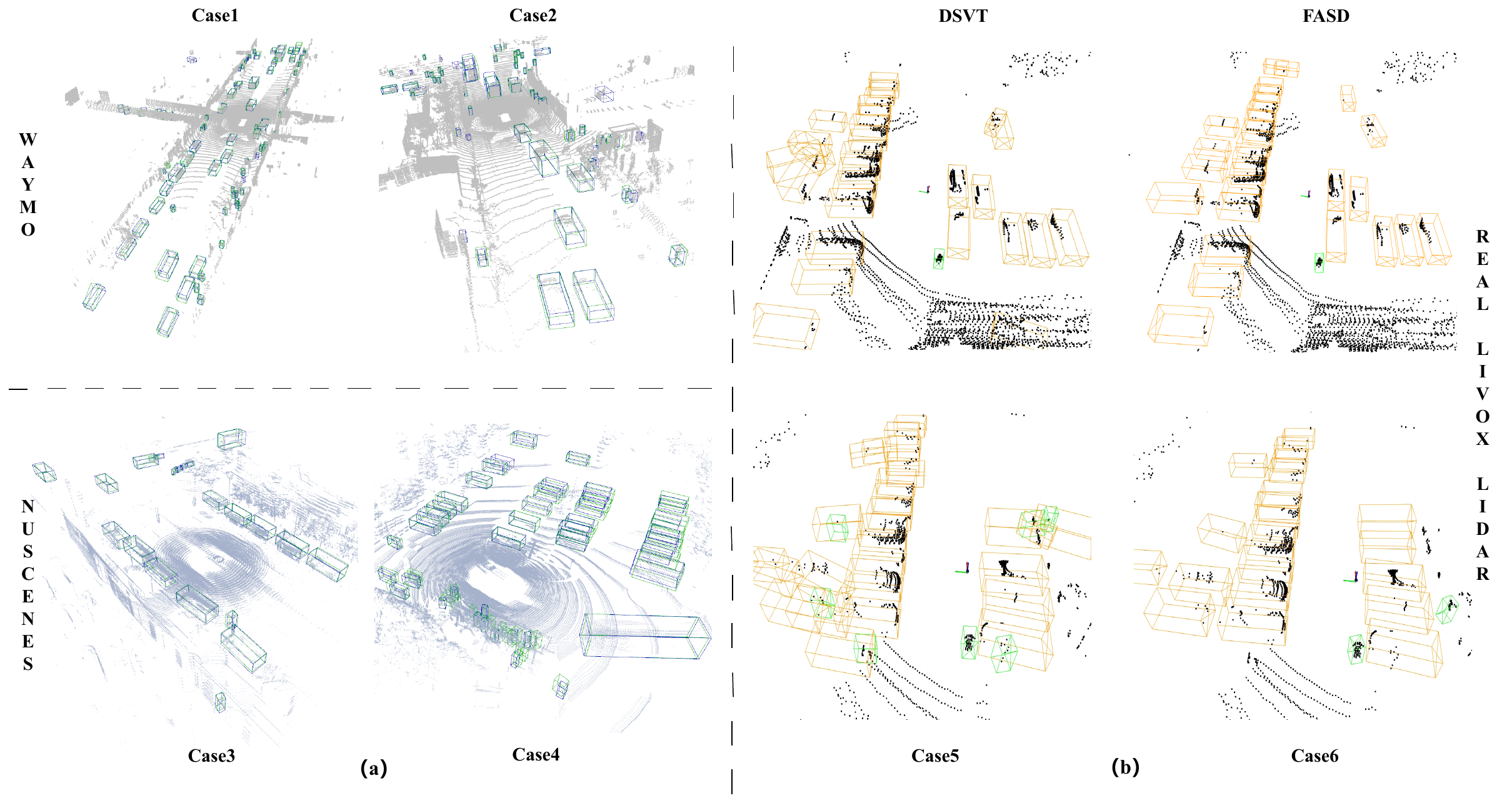}  
    \caption{\textbf{Qualitative visualization} of FASD on the Waymo (gray) and nuScenes (purple). We display predictions (blue) and ground truth (green) in the left image. In the right, boxes indicate the car (orange) and pedestrian (green) for comparison.}
    \vspace{-0.2cm}
    \label{fig5}
\end{figure*}

\begin{figure}
    \centering
    \includegraphics[width=0.48\textwidth]{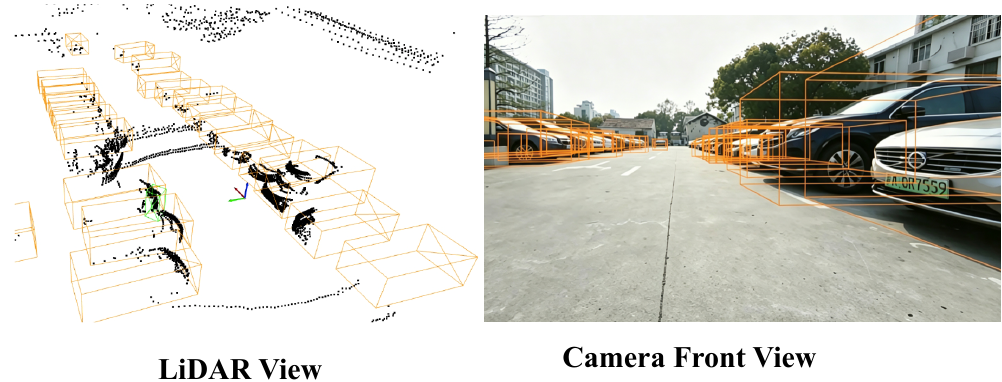}
    \caption{{{Real-World Deployment \textbf{Qualitative Visualization}.}}}
    \label{fig6}
    \vspace{-0.5cm}
\end{figure}

In our experimental setup, we follow the default settings of OpenPCDet \cite{openpcdet} and conduct the experiments using two 24GB Nvidia RTX 3090 GPUs.
We use the AdamW optimizer with {{an}} initial learning rate of \(3 \times 10^{-3}\) and apply layer-wise learning rate decay over 24 epochs.
{{We set the voxel token sequence length $S$ to 512}} in the dynamic voxel group to achieve a longer sequence length and {{a}} larger receptive field, and set {{the number of encoder layers}} $N_L$ to 3.
We first train a separate Transformer-based teacher model and {{use}} \(L_{KD}\) from Eq. (\ref{eq:LKD}) to transfer knowledge to the Mamba-based student model.
Eventually, the hyperparameters \(\lambda_{1}\), \(\lambda_{2}\), and \(\lambda_{3}\) in \(L_{KD}\) are set to 1, 1, and 0.5, respectively.


\subsection{Results and Analysis}

\noindent\textbf\noindent\textbf{Performance on Waymo Dataset.}
We validate the effectiveness of the proposed FASD on the Waymo validation set (Table \ref{tab:1}).
Our teacher model demonstrates a significant improvement over the single-stage DSVT \cite{dsvt}, validating its suitability as a teacher model.
Compared with previous distillation methods \cite{sparsekd, CaKDP}, our student model achieves superior detection performance.
The student model also outperforms Transformer-based DSVT \cite{dsvt} and Mamba-based model \cite{lion, voxelmamba}, indicating stronger contextual and spatial modeling capabilities.
Additionally, it outperforms fully sparse {{detectors}} \cite{voxelnext,safdnet} by 2--5 {{percentage points}} across multiple metrics. 
Our student model not only reduces FLOPs but also significantly improves {{the}} average performance across all categories, with especially strong gains in cyclist detection.
These results {{demonstrate}} that distillation enables the Mamba-based student model to {{capture richer}} global context and spatial awareness, {{thereby}} significantly enhancing {{its}} perception capabilities.

\noindent\textbf\noindent\textbf{Performance on nuScenes Dataset.}
On the nuScenes dataset, FASD outperforms the benchmarks such as Centerpoint \cite{centerpoint} and Voxelnext \cite{voxelnext}, improving NDS and mAP by 2--3 {{percentage points}} compared to DSVT \cite{dsvt}.
Additionally, it surpasses VoxelMamba \cite{voxelmamba} in NDS, demonstrating the effectiveness of heterogeneous distillation in enhancing global contextual representations.
Meanwhile, the improvements in ATE, ASE, and AVE, as shown in Table \ref{tab:2}, indicate more accurate object localization and geometric estimation.

\noindent\textbf{Accuracy–Efficiency Analysis on WOD.}
Under identical voxelization and sparse convolution settings, FASD outperforms SparseKD \cite{sparsekd} and CaKDP \cite{CaKDP} in detection accuracy, as shown in Table \ref{tab:3}.
{{Furthermore, the proposed voxel adapter and distillation strategy achieve 5.92 and 7.76 percentage-point gains in L2 mAPH over CaKDP and SparseKD, respectively, with minimal complexity increase.}}
These results demonstrate that our cross-paradigm distillation framework effectively enhances perception performance while maintaining a favorable accuracy--efficiency trade-off.

\noindent\textbf{Computational Performance Experiment.}
To evaluate real-world deployment efficiency, we compare FASD with representative LiDAR detectors in Table \ref{tab:4} in terms of parameters, FLOPs, memory usage, and inference speed.
For a fair comparison, all methods adopt identical voxelization settings and are evaluated {{on}} LiDAR data collected {{using}} the Livox Mid-360.
Compared with Transformer-based DSVT \cite{dsvt} and our teacher model, the student model achieves a 4\(\times\) reduction in memory usage and nearly 2\(\times\) faster inference, while also surpassing fully sparse detectors \cite{voxelnext,safdnet} in deployment efficiency.
This indicates that Mamba improves computational efficiency through selective scanning and efficient hardware awareness.
Furthermore, compared with VoxelMamba \cite{voxelmamba}, FASD provides a more lightweight architecture with superior real-world efficiency while maintaining competitive detection performance.
Overall, these results demonstrate that heterogeneous knowledge distillation enhances Mamba's contextual and geometric representation capabilities while preserving its deployment efficiency.

\noindent\textbf{Qualitative Visualization.}
As shown in Fig. \ref{fig5}(a), we qualitatively validate our proposed FASD on the Waymo and nuScenes datasets, with green representing the ground truth and blue indicating the predictions.
The results demonstrate that our model {{achieves competitive detection performance}} on both datasets.
As shown in Fig. \ref{fig5}(b), we use LiDAR data from the Livox Mid-360 to compare FASD and DSVT, showing better performance in false negative detection and orientation estimation.
For further quantitative and qualitative analysis, Fig. \ref{fig6} presents representative Front View and BEV View comparisons, with additional results provided in the appendix.

\begin{table}
\centering
\caption{\textbf{Ablation studies of Teacher} on Waymo val set.}
\setlength{\tabcolsep}{1.0mm}
\renewcommand{\arraystretch}{1.05}
\resizebox{1.0\linewidth}{!}{
\begin{tabular}{ccc|cccc}
\hline
\textbf{DVG} & \textbf{VD} & \textbf{AAM}  & \textbf{L1 mAP} & \textbf{L1 mAPH} & \textbf{L2 mAP} & \textbf{L2 mAPH} \\
\hline
\ding{55} & \ding{55} & \ding{55} & 79.02 & 76.62 & 72.96 & 70.56\\
\ding{51} & \ding{55} & \ding{55} & 79.15 & 76.87 & 73.03 & 70.62\\
\ding{51} & \ding{51} & \ding{55} & 80.36 & 77.92 & 74.03 & 71.97 \\
\ding{51} & \ding{51} & \ding{51} & 81.22 & 79.25 & 75.23 & 73.28 \\
\hline
\end{tabular}
}
\label{tab:5}
\vspace{-0.0in}
\end{table}

\begin{table}
\centering
\caption{\textbf{Ablation of KD methods} on Waymo val set. {${L_{feat}^{KD}}$}, {${L_{span}^{KD}}$}, {${L_{logit}^{KD}}$} denote feature, spanhead, logit distillation. {${L_{logit}^{KD}}$} constrains the classification of two models.}
\setlength{\tabcolsep}{1.0mm}
\resizebox{1.0\linewidth}{!}{
\begin{tabular}{c|ccc|ccc}
\hline
\renewcommand{\arraystretch}{5.5}  
\textbf{Adapter} & {$\mathbf{L_{feat}^{KD}}$} & {$\mathbf{L_{span}^{KD}}$} & {$\mathbf{L_{logit}^{KD}}$} & \textbf{Veh. mAP} & \textbf{Ped. mAP} & \textbf{Cyc. mAP} \\
\hline
w/o & \ding{55} & \ding{55} & \ding{55} & 78.86 & 83.05 & 78.13 \\
w/o & \ding{51} & \ding{55} & \ding{55} & 79.54 & 83.45 & 79.03 \\
w/o & \ding{55} & \ding{51} & \ding{55} & 79.21 & 83.94 & 78.77 \\
w/o & \ding{51} & \ding{51} & \ding{55} & {79.36} & {83.76} & {78.90} \\
w/o & \ding{55} & \ding{55} & \ding{51} & 78.09 & 82.87 & 77.04 \\
\hline
w/ & \ding{51} & \ding{55} & \ding{55} & \textbf{80.49} & 84.43 & \textbf{79.56} \\
w/ & \ding{55} & \ding{51} & \ding{55} & 80.01 & \textbf{84.93} & 79.08 \\
w/ & \ding{51} & \ding{51} & \ding{55} & \underline{80.43} & \textbf{84.93} & \underline{79.45} \\
\hline
\end{tabular}
}
\label{tab:5}
\vspace{-0.2cm}
\end{table}


\subsection{Ablation Studies}

\noindent\textbf{Ablation Studies of Teacher Model.}
In the ablation study, we evaluate the fully sparse Transformer teacher model by comparing the effects of voxel diffusion (VD), dynamic voxel grouping (DVG), and adaptive attention mechanism (AAM), with DSVT \cite{dsvt} as the baseline.
The proposed DVG follows the same logic as DSVT, {{with only minor modifications to accommodate the different}} voxel size. 
While VD provides effective key voxel features for full sparse detection, providing a higher baseline.
{{Meanwhile, AAM further enhances the model with adaptive receptive fields through global context and positional information fusion, improving overall performance by approximately 1.2 percentage points across all metrics.}}

\noindent\textbf{Ablation Studies of Knowledge Distillation Methods.}
First, we compare the impact of feature, logit, and span-head distillation on model performance, while layers ablation are shown in the appendix.
By directly distilling intermediate features from Transformer to Mamba, we achieve significant accuracy improvements for large objects (Veh. \& Cyc.).
We argue that the global context information learned through distillation significantly enhances the model's ability to represent large {{objects}}.
Furthermore, the proposed Span-KD improves pedestrian detection by projecting heterogeneous features into a unified {{logit space}}, providing fine-grained semantic supervision and facilitating cross-model knowledge transfer.
In contrast, traditional {{logit}} distillation tends {{to degrade performance}} due to conflicting ground truth information.
Finally, combining the alignment adapter with Span-KD and feature distillation yields the best performance, demonstrating effective transfer of global context to Mamba from both latent and {{logit}} spaces.


{{
\noindent\textbf{Generality verification.}
To verify the generality of FASD, we further integrate our distillation framework with the DSB and IWP designs from VoxelMamba.
As shown in Table~\ref{tab:7}, incorporating VoxelMamba designs (Voxelization, DSB, and IWP) consistently improves the baseline detector, demonstrating their effectiveness.
More importantly, applying FASD to the enhanced architecture further improves Vehicle, Pedestrian, and Cyclist mAP from 80.02\%, 84.15\%, and 79.16\% to 80.97\%, 85.24\%, and 80.05\%, respectively.
These results demonstrate that our knowledge distillation framework is compatible with advanced Mamba architectures and provides complementary gains beyond backbone optimization.
}}

\noindent\textbf{Module Latency Ablation Study.}
Table~\ref{tab:8} presents a module-wise analysis of computational costs
. Although the proposed modules introduce modest computational overhead, the distilled Mamba decoder achieves substantially lower FLOPs and latency than the Transformer decoder while maintaining strong representation capability, demonstrating its effectiveness for real-time LiDAR perception.



\begin{table}[t]
\centering
\caption{\textbf{Ablation studies of VoxelMamba architecture and FASD distillation} on the Waymo validation dataset.}
\setlength{\tabcolsep}{1.0mm}
\resizebox{1.0\linewidth}{!}{
\begin{tabular}{c|ccc|ccc}
\hline
\renewcommand{\arraystretch}{1.0}
\textbf{Voxel} & \textbf{DSB} & \textbf{IWP} & \textbf{FASD} &
\textbf{Veh. mAP} & \textbf{Ped. mAP} & \textbf{Cyc. mAP} \\
\hline
\ding{51} & \ding{55} & \ding{55} & \ding{55} & 78.86 & 83.05 & 78.13 \\
\ding{51} & \ding{51} & \ding{55} & \ding{55} & 79.25 & 83.56 & 78.62 \\
\ding{51} & \ding{51} & \ding{51} & \ding{55} & 80.02 & 84.15 & 79.16 \\
\ding{51} & \ding{51} & \ding{51} & \ding{51} & \textbf{80.97} & \textbf{85.24} & \textbf{80.05} \\
\hline
\end{tabular}
}
\label{tab:7}
\vspace{-0.3cm}
\end{table}

\begin{table}
\centering
\caption{\textbf{Module-wise Computational Cost Analysis}.}
\setlength{\tabcolsep}{1.0mm}
\renewcommand{\arraystretch}{1.05}
\resizebox{1.0\linewidth}{!}{
\begin{tabular}{cccccc|cc}
\hline
\textbf{Voxel} & \textbf{Conv} & \textbf{DVG} & \textbf{VD}& \textbf{Transformer} & \textbf{Mamba} & \textbf{Flops} & \textbf{Latency} \\
\hline
\ding{51} & \ding{55} & \ding{55} & \ding{55} & \ding{55} & \ding{55} & 0.0 G & 3.8 ms\\
\ding{51} & \ding{51} & \ding{55} & \ding{55} & \ding{55} & \ding{55} & 21.3 G & 25.4 ms\\
\ding{51} & \ding{51} & \ding{51} & \ding{55} & \ding{55} & \ding{55} & 29.1 G & 32.8 ms \\
\ding{51} & \ding{51} & \ding{51} & \ding{51} & \ding{55} & \ding{55} & 42.6 G & 42.7 ms \\
\ding{51} & \ding{51} & \ding{51} & \ding{51} & \ding{51} & \ding{55} & 118.3 G & 100.9 ms \\
\ding{51} & \ding{51} & \ding{51} & \ding{51} & \ding{55} & \ding{51} & 53.1 G & 51.3 ms \\
\hline
\end{tabular}
}
\label{tab:8}
\vspace{-0.1in}
\end{table}

\section{CONCLUSIONS}
To enhance the real-time capability of the LiDAR detector, we first implement a robust teacher model via dynamic voxel group and adaptive attention. A high-performance, high-accuracy Mamba-based model is constructed by multi-stage distillation of the heterogeneous model, demonstrating performance improvements on Waymo and {nuScenes} datasets.

In future work, we plan to extend the heterogeneous model {{distillation framework proposed in this paper to other tasks, such as lane perception and end-to-end planning}}.
Additionally, Mamba is primarily used as a student model in this paper, and exploring additional foundation model permutations could help verify performance.
Ultimately, we aim to refine the {{cross-model distillation}} to make it more {{robust and adaptable across diverse domains.}}


\bibliographystyle{IEEEtran}
\bibliography{reference}

\end{document}